\documentclass[12pt]{article}
\usepackage{amsmath}
\usepackage{amssymb}
\usepackage{amsfonts}
\usepackage{epsfig}
\usepackage{mathrsfs}
\usepackage{amsmath,amsthm,amssymb,amscd}
\usepackage{MnSymbol}
\usepackage{rotating}
\usepackage{multirow}
\usepackage{float}
\usepackage[english]{babel}
\usepackage{floatpag}
\usepackage{enumerate}
\usepackage{threeparttable}
\usepackage{color}
\usepackage{tabularx}
\usepackage{booktabs}
\usepackage{lscape}
\usepackage{algorithm}
\usepackage{algorithmic}
\usepackage{makecell}
\usepackage{tabularray}
\usepackage{graphicx}
\usepackage{tabularray}
\usepackage{subcaption}
\UseTblrLibrary{booktabs}
\usepackage{hyperref}
\hypersetup{
hidelinks,
colorlinks=true,
linkcolor=blue,
citecolor=blue,
urlcolor = blue
}
\usepackage[numbers]{natbib}
\usepackage{url}
\usepackage{tikz}

\newcommand\appendix@section[1]{\refstepcounter{section}\orig@section*{#1}\addcontentsline{toc}{section}{#1}}
\let\orig@section\section\g@addto@macro\appendix{\let\section\appendix@section}
\makeatother \makeatletter
\renewcommand\footnoterule{\kern-3\p@ \hrule width 1\columnwidth \kern 2.6\p@}%Footnote Line
\newskip\@footindent
\renewcommand\@footindent{0pt}%Footnote indent
\@addtoreset{footnote}{page}
\long\def\@makefntext#1{\@setpar{\@@par\@tempdima \hsize
\advance\@tempdima-\@footindent \parshape \@ne \@footindent
\@tempdima}\par \noindent \hbox to
\z@{\hss\@thefnmark£º\hspace{0.2em}}#1}

\def\@makefnmark{\hbox{\textsuperscript{\@thefnmark}}}
\makeatother

\newtheorem{thm}{Theorem}
\newtheorem{cor}{Corollary}
\newtheorem{lem}{Lemma}

\theoremstyle{remark}
\newtheorem{assumption}{Assumption}

\title{Deep non-parametric logistic model with case-control data and external summary information}
\author{ \vspace{5mm} Hengchao Shi \ \ Ming Zheng \ \ Wen Yu$^\ast$\\
 Department of Statistics and Data Science \\
 School of Management\\
 Fudan University\\
 \vspace{5mm}
 Shanghai 200433, P.R.China\\
 \vspace{1cm} 
{\small $^\ast$Corresponding author: \ wenyu@fudan.edu.cn}}
\date{ }

\begin{document}
\maketitle
\newpage

\begin{abstract}
The case-control sampling design serves as a pivotal strategy in mitigating the imbalanced structure observed in binary data. We consider the estimation of a non-parametric logistic model with the case-control data supplemented by external summary information. The incorporation of external summary information ensures the identifiability of the model. We propose a two-step estimation procedure. In the first step, the external information is utilized to estimate the marginal case proportion. In the second step, the estimated proportion is used to construct a weighted objective function for parameter training. A deep neural network architecture is employed for functional approximation. We further derive the non-asymptotic error bound of the proposed estimator. Following this the convergence rate is obtained and is shown to reach the optimal speed of the non-parametric regression estimation. Simulation studies are conducted to evaluate the theoretical findings of the proposed method. A real data example is analyzed for illustration.
\end{abstract}

\vfill \hrule \vskip 6pt \noindent {\em MSC:} 62D05 \ 62J12\\
\noindent {\em Some key words}: Case-control sampling; Data integration; Multiple-layer perceptron;  Non-asymptotic error bound; Non-parametric regression.

\newpage

\section{Introduction}\label{intro}

Binary data is prevalent across numerous applications in various fields. When covariate information is available, people often need to address classification problem for the response based on the covariates, or to estimate the association pattern between the response and the covariates. Let $Y$ be the binary response variable, taking values $0$ and $1$, and $X$ the $p$-dimensional covariates vector. The most widely used statistical model for handling the binary response $Y$ is the linear logistic regression model, expressed as follows:
\begin{eqnarray*}
\mathsf{P}(Y=1 \ | \ X)=\frac{\exp(\alpha+\beta^\top X)}{1+\exp(\alpha+\beta^\top X)},
\end{eqnarray*}
where $\alpha$ is the intercept parameter and $\beta$ is the $p$-dimensional slope parameter. It is well known that based on independent and identically distributed (i.i.d.) data, the maximum likelihood method can be employed to obtain efficient estimators for the regression parameters.

An often observed phenomenon in binary data is the presence of the imbalanced data structure. That is, data points belonging to one category are significantly rarer compared to those in the opposite category. Without loss of generality, we use $Y=1$ to represent the rare category, commonly referred to as the case, while the opposite category, denoted as $Y=0$, is called the control. The imbalanced data structure poses challenges for data analysis. For example, when the number of cases in a dataset is too small, the maximum likelihood estimator (MLE) for the linear logistic model may become highly unstable. To mitigate such imbalanced structure, the case-control sampling design \cite{mantel1959statistical, miettinen1976estimability} is one of the most commonly used approach. This sampling design involves drawing separate random samples from the cases and controls, and usually comparable sizes of cases and controls are collected, thereby creating a more balanced sample for analysis. 

Essentially the case-control sampling results in an artificial biased sample. The sampling bias needs to be addressed when applying analysis methods designed for i.i.d. full data to the case-control data. It is well known that if one fits the linear logistic model to the case-control sample and directly obtains the MLE, the MLE for the slope parameter $\beta$ remains consistent, but the MLE for the intercept $\alpha$ is biased. This arises from the fact that the intercept can not be identified from the case-control data \cite{farewell1979some, prentice1979logistic}. Consequently, the marginal case proportion is not estimable based on a single case-control sample either. 

The primary reason that the intercept $\alpha$ is not identifiable is that the score function of $\alpha$ lies within the linear space spanned by the score function of the density of the covariates $X$. It is not difficult to see that the distribution of $X$ is not identifiable from the case-control sampling data. Therefore, if there exists some additional data resource that aids in identifying the marginal covariate distribution, it becomes feasible to recover the intercept $\alpha$ from case-control data. Several authors have made contributions on this aspect. \cite{qin2015using} sought to increase the estimation efficiency of case-control logistic regression by utilizing auxiliary covariate-specific information available from other data resource. \cite{zhang2022integrative} developed an integrative analysis that combines multiple case-control studies with summarized external information. \cite{tang2021combining} found that the marginal distribution of covariates can be identified through different case-control studies even when the studies have varying analysis purpose. \cite{quan2024efficient} considered a semi-supervised framework, which includes a labeled data set drawn by the case-control sampling and an unlabeled data set drawn from the marginal distribution of covariates. The unlabeled covariate data is used to estimate the covariates distribution, thereby making the intercept become estimable. \cite{quan2023semisupervised} discussed semi-supervised inference for case-control data under possibly mis-specified linear logistic model. \cite{shi2024statistical} further demonstrated that the intercept can be identified as long as certain summary information of the covariates is obtained from external data sources. They employed an empirical likelihood approach to estimate the parameters of the linear logistic model.

With recent advancement in data collection capabilities and development on machine learning techniques, many traditional linear models have been extended to non-linear structures through various non-parametric approaches. Similar to \cite{quan2024efficient}, \cite{wang2023semi} considered the case-control data under a semi-supervised framework. They extended the linear logistic regression model in \cite{quan2024efficient} to a non-parametric version. Utilizing the unlabeled covariate data, they provided the model identifiability and adopted a two-stage sieve method for model estimation. Motivated by \cite{wang2023semi}, we also consider non-parametric logistic model in the presence of a data set with binary labels drawn by the case-control sampling and certain unlabeled covariate information. However, we find out that in order to identify the model, it is unnecessary to know the individual-level information of the unlabeled covariate data, as required in \cite{wang2023semi}. It suffices to obtain certain external summary information on covariates. Based on this external summary information of covariates, we propose a two-step estimation procedure for the nonparametric logistic model. In the first step, an estimating equation is developed to estimate the marginal case proportional by utilizing the external summary information. In the second step, an objective function based on the inverse probability weighting technique is developed. For functional approximation, we adopt the multiple-layer perceptrons (MLP). This deep-structured neural network is useful for alleviating curse of dimensionality. We further provide theoretical guarantees for the proposed estimating procedure. The non-asymptotic error bound for the estimation error is derived. Then the consistency of the proposed estimator is established and the convergence rate attaining the optimal speed of non-parametric regression estimation is obtained. To the best of our knowledge, no existing literature has obtained similar theoretical results for neural network based logistic model under case-control sampling.

The rest of the paper is organized as follows. In Section \ref{method}, we first give out the necessary notation and model specification. The identifiablity issue is discussed. Then we present the MLP structure for functional approximation. The proposed two-step estimation procedure is described in details. In Section \ref{thresu}, the theoretical properties of the proposed estimator are provided, including the non-asymptotic error bound and the convergence rate. In Section \ref{numer}, the simulation results and the real data analysis results are presented. Section \ref{conclu} concludes. All the technical details are summarized in the Appendix.

\section{Method}\label{method}
\subsection{Model specification and identifiability} 

We consider the following non-parametric logistic model
\begin{eqnarray}\label{nplogistic}
\mathsf{P}(Y=1 \ | \ X)=\frac{\exp(g(X))}{1+\exp(g(X))},
\end{eqnarray}
where $g(\cdot)$ is an unspecified smooth function to be estimated. Let $F(x)$ be the cumulative distribution function of $X$ and without loss of generality, assume that $F(x)$ has a density denoted by $f(x)$. The primary data set for analysis consists of a random sample of size $n_1$ drawn from the case population (the sub-population with $Y=1$) and a random sample of size $n_0$ from the control population (the sub-population with $Y=0$). The total size is then $n_1+n_0$, denoted by $n$. The primary data set with binary labels is denoted by $\{(y_i,x_i), i=1,\ldots,n\}$.

Use $f_1(x)$ and $f_0(x)$ to denote the conditional density of $X$ given $Y=1$ and $Y=0$, respectively. From the Bayesian rule, it is easy to derive that under the model (\ref{nplogistic}), we have that
\begin{eqnarray}\label{bayesrule}
f_1(x)=\exp\left(\gamma+g(x)\right)f_0(x),
\end{eqnarray}
where $\gamma=\log(\mathsf{P}(Y=0)/\mathsf{P}(Y=1))$. Denote the marginal case proportion $\mathsf{P}(Y=1)$ by $P_1$. Since $P_1$ is not identifiable from a single case-control data, the function $g(\cdot)$ in (\ref{nplogistic}) is not identifiable, either. Strictly speaking, the function can be identified up to the constant $\gamma$. Thus, based on the primary data set, one cannot estimate the function $g(\cdot)$ consistently.

Suppose that beside the primary data set, one can also obtain some summary-level information on the covariate distribution $F$. For instance, let $h(\cdot)$ be a known function from $\mathbb{R}^p$ to $\mathbb{R}$. Define $\mu=\mathsf{E}(h(X))$. If one could know the value of $\mu$ from some external data sources, from the law of total expectation, we have that
\begin{eqnarray}\label{totexp}
P_1\mu_1+(1-P_1)\mu_0=\mu,
\end{eqnarray}
where $\mu_1$ and $\mu_0$ are the conditional expectation of $h(X)$ given $Y=1$ and $Y=0$, respectively, that is, $\mu_1=\int_{\cal X}h(x)f_1(x)dx$ and $\mu_0=\int_{\cal X}h(x)f_0(x)dx$, with $\cal X$ being the support of $X$. Since $\mu_1$ and $\mu_0$ can be identified from the case-control sample as long as $n_1$ and $n_0$ goes to infinity, $P_1$ is identifiable from (\ref{totexp}). Consequently, both $\gamma$ in (\ref{bayesrule}) and $g(\cdot)$ in (\ref{nplogistic}) can be identified.

Note that even when the true value of $\mu$ is not available, the equation (\ref{totexp}) is still useful as long as some consistent estimate of $\mu$, denoted by $\tilde{\mu}$ in the rest of the paper, can be obtained from external data sources. For instance, in the semi-supervised framework, the unlabeled data consists of a random sample from $F(x)$, denoted by $\{x_i, i=1,\ldots,n_e\}$. One can set $\tilde{\mu}=n_e^{-1}\sum_{i=1}^{n_e}h(x_i)$. However, the proposed method does not require the individual-level information of the external data set. Only an estimator of $\mu$ would be adequate.

\subsection{Functional approximation}

To approximate the unspecified function $g(\cdot)$, the MLP is used. Let $D$ be the depth of the MLP, ${\bf W}=\{W_i\}_{i=0}^D$ be the set of weight matrices in $D+1$ layers with $W_i\in\mathbb{R}^{p_{i+1}\times p_i}$, and ${\bf b}=\{b_i\}_{i=0}^D$ be the set of bias vectors in $D+1$ layers with $b_i\in\mathbb{R}^{p_{i+1}}$. We use the widely used ReLU as the activation function, denoted by $\sigma(x)=\max\{0,x\}$. Then for the interested function $g(x)$, the corresponding MLP, denoted by $\widehat{g}(x;{\bf W},{\bf b})$, is given by
\begin{eqnarray*}
\widehat{g}(x;{\bf W},{\bf b})=W_D\sigma\left(\cdots\sigma\left(W_1\sigma\left(W_0x+b_0\right)+b_1\right)+\cdots\right)+b_D.
\end{eqnarray*}
Let $W=\max\{p_1,\ldots,p_D\}$ be the width of $\widehat{g}(x;{\bf W},{\bf b})$ and $S=\sum_{i=0}^Dp_{i+1}\times(p_i+1)$ be the parameter size, i.e., the total number of parameters of $\widehat{g}(x;{\bf W},{\bf b})$. It has been showed in \cite{bartlett2019nearly, jiao2023deep} that the depth, the width, and the size satisfy the following relationship
\begin{eqnarray}\label{f3}
\max\{W,D\}\leqslant S \leqslant W(d+1) + (W^{2} + W)(D-1) + W + 1 = O(W^{2}D).
\end{eqnarray}
Such fully-connected neural network have been widely used in statistical learning literature; see \cite{neyshabur2017pac, elbrachter2021deep, zhong2022deep}.

\subsection{Model estimation} 

For ease of presentation, write the Sigmoid function as $\psi(x)$, i.e., $\psi(x)=\exp(x)/(1+\exp(x))$. We propose a two-step estimation procedure to estimate the model (\ref{nplogistic}). The main idea is described as follows. For $n$ i.i.d. data, the negative log-likelihood function based on (\ref{numer}) is given by
\begin{eqnarray*}
l_n=-\frac{1}{n}\sum_{i=1}^n\left[y_i\log\psi(g(x_i))+(1-y_i)\log\left(1-\psi(g(x_i)\right)\right].
\end{eqnarray*}
However, for the case-control sample, using $l_n$ directly results in biased estimation. To avoid such bias, we consider using the inverse probability weighting approach, that is, adjusting $l_n$ to
\begin{eqnarray}\label{theobj}
l_n^w=-\frac{1}{n}\sum_{i=1}^n\left[w_1^{-1}y_i\log\psi(g(x_i))+w_0^{-1}(1-y_i)\log\left(1-\psi(g(x_i)\right)\right],
\end{eqnarray}
where $w_1=n_1/(nP_1)$ and $w_0=n_0/[n(1-P_1)]$. Note that $w_1$ is proportional to the sampling probability of the cases from the population, while $w_0$ is proportional to the sampling probability of the controls. It is not difficulty to show that the limit of $l_n^w$ as $n\to\infty$ equals to $-\mathsf{E}[Y\log\psi(g(X))+(1-Y)\log(1-\psi(g(X)))]$, which is also the limit of $l_n$ for i.i.d. data.

To implement the main idea, in the first step we estimate the weights in $l_n^w$ from the equation (\ref{totexp}). Note that $\mu_1$ and $\mu_0$ can be naturally estimated from the case-control sample. Specifically, we use $\hat{\mu}_1=n_1^{-1}\sum_{i=1}^{n}y_ih(x_i)$ and $\hat{\mu}_0=n_1^{-1}\sum_{i=1}^{n}(1-y_i)h(x_i)$ to estimate $\mu_1$ and $\mu_0$, respectively. From the equation (\ref{totexp}), it is easy to obtain an estimator of $P_1$, defined as $\hat{P}_1=(\tilde{\mu}-\hat{\mu}_0)/(\hat{\mu}_1-\hat{\mu}_0)$. Thus, the weights in $l_n^w$ can be estimated by $\hat{w}_1=n_1/(n\hat{P}_1)$ and $\hat{w}_0=n_0/[n(1-\hat{P}_1)]$.

Based on the estimated weights, in the second step we propose the following objective function
\begin{eqnarray}\label{obj}
l_n^w({\bf W},{\bf b})=-\frac{1}{n}\sum_{i=1}^n\left[\hat{w}_1^{-1}y_i\log\psi(\widehat{g}(x_i;{\bf W},{\bf b}))+\hat{w}_0^{-1}(1-y_i)\log\left(1-\psi(\widehat{g}(x_i;{\bf W},{\bf b})\right)\right].
\end{eqnarray}
We minimize $l_n^w({\bf W},{\bf b})$ over $({\bf W},{\bf b})$ and denote the minimizer by $(\widehat{\bf W}_n,\widehat{\bf b}_n)$. Then the neural network based estimator of $g(x)$ is defined as $\widehat{g}_n(x)=\widehat{g}(x;\widehat{\bf W}_n,\widehat{\bf b}_n)$. 

It is worth mentioning that the weights in the proposed objective function do not contain the neural network parameters, making the objective function convex in the parameters. So it is easy to apply the stochastic gradient descent (SGD) based optimizer (\cite{kingma2014adam, ruder2016overview}) to minimize the objective function (\ref{obj}). %The details of the minimization algorithm are put in the Appendix.

\section{Theoretical results}\label{thresu}

In this section, we first show the asymptotic properties of the marginal case proportion estimator $\hat{P}_1$. Then a non-asymptotic error bound of the excess risk of the empirical risk minimizer (ERM) is establihed and the convergence rate of $\widehat{g}_n(x)$ is derived. For $\beta\in\mathbb{N}^+$ (positive integers) and a constant $B>0$, define the following Sobolev space on the support of covariates 
\begin{eqnarray*}\label{f12}
\mathcal{W}_B^\beta (\mathcal{X}) = \left\{f:{\underset{\alpha,||\alpha ||_1 \leqslant \beta}{\max}} \underset{x \in \mathcal{X}}{\text{esssup}} |D^\alpha f(x)| \leqslant B  \right\}
\end{eqnarray*}
where $\alpha_i \in {\mathbb N}_0$ (non negative integers), $i=1, ..., p$, ${\alpha} = (\alpha_1, \ldots, \alpha_p)$, $\left|\mathbf{\alpha}\right| = \sum_{i=1}^p\alpha_i$, and $D^{\mathbf{\alpha}}f =(\partial^{\left|\mathbf{\alpha}\right|}f)/(\partial x_1^{\alpha_1} \cdots\partial x_d^{\alpha_d})$ is the weak derivative of $f$. Some assumptions are required.

\begin{assumption}\label{ass1}
    The support $\cal X$ is a compact set in $\mathbb R^d$ and there exists some constant $c_1>1$ such that $\sup_{x\in{\cal X}}f(x)\leqslant c_1$.
\end{assumption}

\begin{assumption}\label{ass2}
    The true value of the function $g(\cdot)$ lie in $\mathcal{W}_B^\beta(\mathcal{X})$ for some $\beta\in\mathbb{N}^+$ and constant $B>1$. 
\end{assumption}

\begin{assumption}\label{ass3}
    The space ${\cal G}_n$ is constructed as a set of MLPs satisfying $\|g_n\|_{\infty}\leqslant B$ for all $g_n\in{\cal G}_n$ with depth $D_n$, width $W_n$, and parameter size $S_n$.
\end{assumption}

\begin{assumption}\label{ass4}
    $h(X)$ is bounded almost surely and $\mu_1\not=\mu_0$.
\end{assumption}

\begin{assumption}\label{ass5}
    $\tilde{\mu}$ satisfies that $\sqrt{n}(\tilde{\mu}-\mu)\xrightarrow{d} N(0,V)$ as $n\to\infty$, where $\xrightarrow{d}$ stands for converging in distribution and $V>0$.
\end{assumption}

\noindent{\it Assumption \ref{ass1}} is a standard regularity condition for covariates distribution. {\it Assumption \ref{ass2}} imposes certain smoothness on the underlying function in model ({\ref{nplogistic}). {\it Assumption \ref{ass3}} aims to construct the space of functional approximation by restricting the complexity of the MLPs to scale with the primary data size $n$. {\it Assumption \ref{ass4}} guarantees that the marginal case proportion can be identified from the equation (\ref{totexp}). It is satisfied as long as $g(x)$ is not degenerate. {\it Assumption \ref{ass5}} can be satisfied in the applications where the external data size is comparable to the primary data size and the estimator $\tilde{\mu}$ has normal statistical properties.  

We first give out the consistency and asymptotic normality of the marginal case proportion estimator $\hat{P}_1$ in the following theorem, which is proved in the Appendix. 

\begin{thm}\label{thm1}
Under {\it Assumption 4} and {\it Assumption 5}, we have that i) $\hat{P}_1\xrightarrow{p}P_1$ as $n\to\infty$, where $\xrightarrow{p}$ stands for converging in probability, and ii) $\sqrt{n}(\hat{P}_1 - P_1) \xrightarrow{d} N(0,a_1^2 V_1 + a_2^2 V_0 + a_3^2 V)$ as $n\to\infty$, where $a_1, a_2, a_3, V_0$, and $V_1$ are given in the Appendix.
\end{thm}

To present the non-asymptotic error bound, we need some more notation. Let $Z=(Y,X)$ and $l(Z;g)=\hat{w}_1^{-1}Y\log\psi(g(X))+\hat{w}_0^{-1}(1-Y)\log(1-\psi(g(X)))$. Define $r(Z;\widehat{g}_n)=l(Z;\widehat{g}_n)-l(Z;g)$, where $\widehat{g}_n$ stands for the proposed estimator $\widehat{g}_n(x)$. Let $VCdim(\mathcal{G}_n)$ and $Pdim(\mathcal{G}_n)$ be the VC-dimension and pseudo dimension (see the detailed definition in \cite{jiao2023deep}) of ${\cal G}_n$. Write ${\cal S}=\{(y_i,x_i), i=1,\ldots,n\}$ to be primary data set. The following theorem, proved in the Appendix, gives out the non-asymptotic error bound of the excess risk of the ERM.

\begin{thm}\label{thm2}
Suppose that $\mathcal{G}_n$ consists of MLPs with width $W_n = 38\beta^2N_n\lceil\log_2(8N_n)\rceil$ and depth $D_n = 21\beta^2M_n\lceil\log_2(8M_n)\rceil$ for some $M_n$ and  $N_n \in \mathbb{N}^+$. Under {\it Assumption \ref{ass1}} to {\it Assumption \ref{ass5}}, for $n > Pdim(\mathcal{G}_n)/2$, we have that
\begin{eqnarray*}
\mathsf{E}_\mathcal{S} \mathsf{E}_Z\left[r(Z;\hat{g}_n)\right]\leqslant c_2\left(2\hat{w}_1^{-1} +\hat{w}_0^{-1}\right)^2 B^2\frac{S_nD_n\log S_n \log n}{n} + 18\left(2\hat{w}_1^{-1} +\hat{w}_0^{-1}\right)B\beta^2 p^{3/2}(N_nM_n)^{-2\beta/p},
\end{eqnarray*}
where $c_2$ is a positive constant.
\end{thm}

With proper choice of the depth of MLP, we can derive the following convergence rate of the proposed estimator $\widehat{g}_n(x)$.

\begin{cor}\label{cor1}
Suppose that $\mathcal{G}_n$ consists of MLPs with width $W_n = 38\beta^2N_n\lceil\log_2(8N_n)\rceil$ and depth $D_n = 21\beta^2M_n\lceil\log_2(8M_n)\rceil$. Let $N_n$ be fixed integer and $M_n=O(n^{p/2(2\beta+p)})$. Under {\it Assumption 1} to {\it Assumption 5}, we have that
\begin{eqnarray*}
\mathsf{E}_X\left|\widehat{g}_n(X)-g(X)\right|=O_p\left(n^{-\beta/(2\beta+p)}\right).
\end{eqnarray*} 	
\end{cor}
	
\noindent Note that the dominated term given in Corollary \ref{cor1} reaches the optimal convergence rate in the classical non-parametric regression \cite{stone1982optimal}. Here we choose the depth of the MLP to grow with the sample size while keeping the width fixed. The same convergence rate may be attained by growing the width and keeping the depth fixed. However, following the finding of exiting literature \cite{jiao2023deep}, deep networks with fixed width require smaller parameter size than the wide ones with fixed depth to achieve the same convergence rate.

\section{Numerical results}\label{numer}

Extensive simulation studies have been conducted to assess the theoretical findings established in Section \ref{thresu} and we report some of them for illustration. We also set up a situation consisting of a case-control sample and external data information from a real data example. The performance of the proposed method is illustrated by the real data example.

\subsection{Simulation studies}

We start with the univariate covariate case, i.e., $p=1$. The covariate $X$ is generated from a uniform distribution on $[0,2]$. Four different types of $g(x)$ are considered and the specific forms are given in Table \ref{table1}. The corresponding marginal case proportions and the value of $\gamma$ are also listed. For the primary case-control data set, we set $n_1=n_0=500$. A random sample of the covariate with size $2000$ is drawn independently to generate external summary information. We use $h(x)=x$, that is, the external summary information used is the mean of the covariate. For the functional approximation, we train an MLP by SGD, with the learning rate being set to be 0.01. The training process is repeated for 10,000 epochs, until the loss change being less than the predetermined criteria. We use $80\%$ of the primary data set as the training set and the remaining $20\%$ as the validation set for tuning the hyperparameters of the network, including the depth and the width. We tune $D$ to be 2, 3 or 4 and $W$ to be 64 or 128. The best configuration is chosen according to the prediction accuracy on this validation set. Then the proposed estimator $\widehat{g}_n(x)$ is obtained. We also calculate the neural network based estimator without using the external information, denoted by $\widetilde{g}_n(x)$, for the purpose of comparison. Obviously, $\widetilde{g}_n(x)$ is biased with the magnitude of $\gamma$. One hundred replications are carried out. In Figure \ref{fig1}, we plot the average of the two estimators against the true function value on the covariate support under four different types. We can see that $\widehat{g}_n(x)$ approximates the true function quite well. $\widetilde{g}_n(x)$ is able to capture the functional shape, but there exists a significant shift due to the bias brought by the case-control sampling.

\begin{table}[h]
  \centering
  \caption{Specific forms of $g(x)$ in univariate case}
  \begin{tabular}{ccc}
    \toprule
    Type & $g_0(x)$ & $p$ \\
    \midrule
    $T_1$   & $-3 + 2x$   & 0.316   \\
    % $S_2$   & $-5 + e^x$   & 0.237   \\
    $T_2$   & $-2 + 3\sin(4x)$   & 0.312   \\
    % $S_3$   & $ -8 -3x + x^2$   & 0.256   \\
    $T_3$   & $ -2(x-1)^2$   & 0.352   \\
    $T_4$   & $-2 - \log{x}$   & 0.187   \\
    \bottomrule
  \end{tabular}
  \label{table1}
\end{table}

\begin{figure}[htbp]
    \centering
    \includegraphics[width=\textwidth]{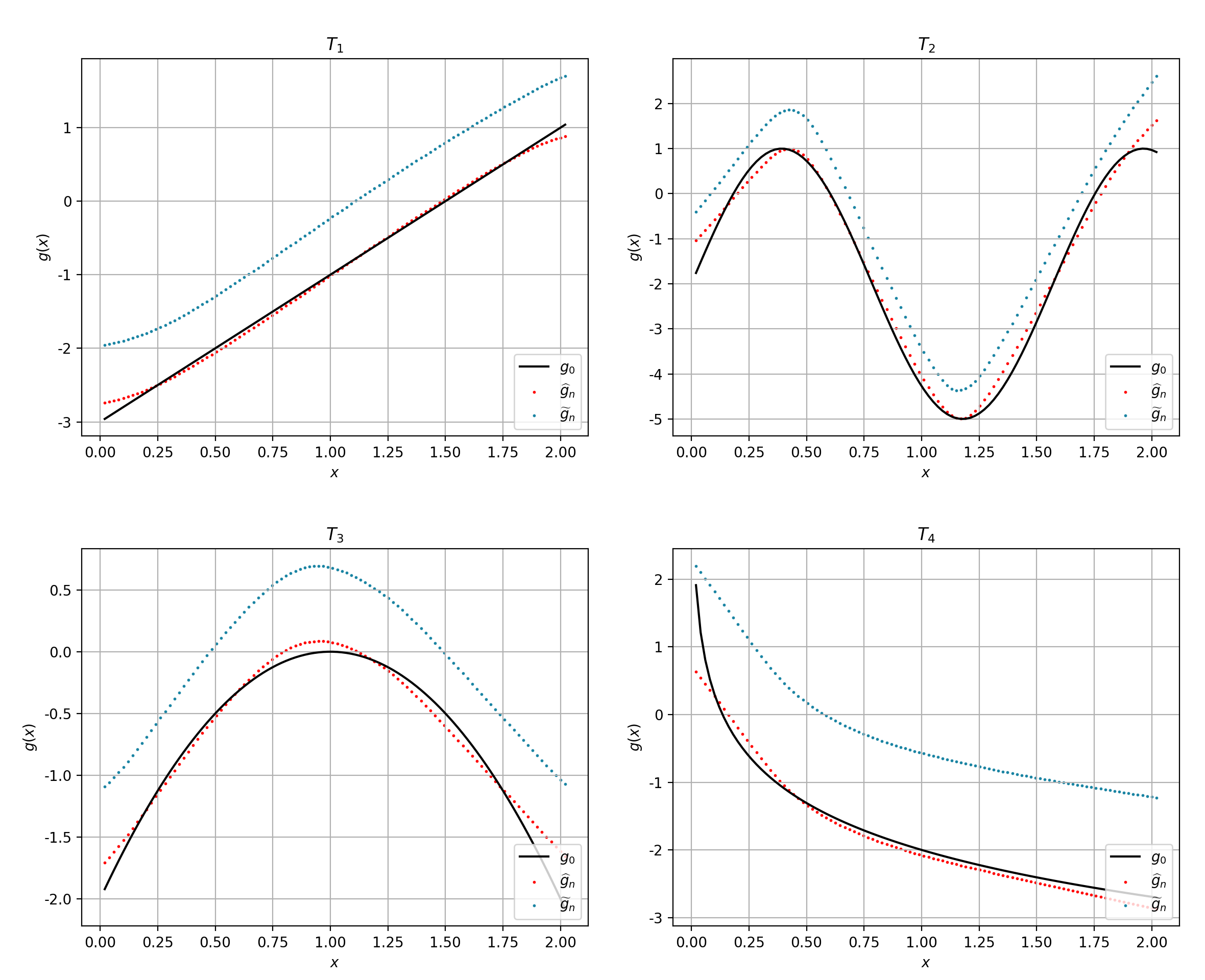}
    \caption{The average of $\widehat{g}_n(x)$ and $\widetilde{g}_n(x)$. The black line corresponds to the truce value of $g(x)$; the red dotted line corresponds to the average of $\widehat{g}_n(x)$; the blue dotted line corresponds to the average of $\widetilde{g}_n(x)$}\label{fig1}
    % \begin{tablenotes}
    %     \item  Note: 
    % \end{tablenotes}
\end{figure}

For the multiple covariates case, we set $p=6$, i.e., $X=(X_1,\ldots,X_6)$. All the covariates are generated from the uniform distribution on $[0,2]$ independently. Two types of $g(x)$ are considered and the specific forms are given in Table \ref{tabel2}. For the primary case-control data set, we consider three different case and control sizes. In the first one, $n_1=n_0=1000$. In the second one, $n_1=500$ and $n_0=1500$. In the third one, $n_1=200$ and $n_0=1800$. We use $60\%$ of the data as the training set, $20\%$ as the validation set, and the remaining $20\%$ as the testing set for evaluation. A random sample of the covariate with size 2000 is drawn independently to external information. We use $h(x)=x_1$, i.e., the mean of $X_1$, to be the external summary information. For functional approximation, we use the same training and validation procedure as those in the univariate case. To evaluate the performance, for any estimator denoted by $\hat{g}$, we define the following relative error on the testing set
\begin{eqnarray*}
\mathsf{RE}(\hat{g})=\frac{\sum_{i\in{\cal T}}|\hat{g}(x_i)-g(x_i)|}{\sum_{i\in{\cal T}}|g(x_i)|},
\end{eqnarray*} 
where $\cal T$ is the index set of the testing set. One hundred replications are carried out. We compare the average of $\mathsf{RE}(\widehat{g}_n)$ and $\mathsf{RE}(\widetilde{g}_n)$ in various setups. The results are summarized in Table \ref{tabel3}.

From the results, we can see that in almost all the steps, the average relative error of $\widehat{g}_n$ is significantly smaller that that of $\widetilde{g}_n$. The difference is more obvious for the second form of $g(x)$. The proposed method effectively reduces the estimation bias in case-control data by utilizing external summary information.

\begin{table}[h]
  \centering
  \caption{Specific forms of $g(x)$ in multivariate case}
  \begin{tabular}{ccc}
    \toprule
    Type & $g_0(x)$ & $p$ \\
    \midrule
    $T_5$   & $-4 x_1 + x_2 + x_3 -3 x_4^2 + \sqrt{x_5 \exp(x_6)}$   & 0.188   \\
    $T_6$   & $\sin(x_1 + x_2) + x_3 -2 x_4^2 + x_5\log(x_6 +3)$   & 0.375   \\
    \bottomrule
  \end{tabular}
  \label{tabel2}
\end{table}

\begin{table}[h]
    \centering
    \caption{The relative error of the estimators in multivariate case}
    \begin{tabular}{ccccc}
        \toprule
        Type & $n_0$ &$n_1$& $\mathsf{RE}(\widehat{g}_n)$  & $\mathsf{RE}(\widetilde{g}_n)$ \\
        \midrule
        \multirow{3}{*}{$T_5$} 
          & 1000 &1000& 0.2110 \scriptsize{(0.0261)} &  0.4956 \scriptsize{(0.0441)}   \\
          & 1500&500  & 0.1956 \scriptsize{(0.0239)} & 0.2222 \scriptsize{(0.0318)} \\
          & 1800&200  & 0.2094 \scriptsize{(0.0385)} & 0.2324 \scriptsize{(0.0369)} \\
        \midrule
        \multirow{3}{*}{$T_6$} 
          & 1000 &1000 & 0.2140 \scriptsize{(0.0366)} & 0.6546 \scriptsize{(0.0490)} \\
          & 1500&500  & 0.1817 \scriptsize{(0.0256)} & 0.2285 \scriptsize{(0.0344)} \\
          & 1800&200  & 0.1925 \scriptsize{(0.0336)} & 0.2062 \scriptsize{(0.0421)} \\
        \bottomrule
    \end{tabular}
    \label{tabel3}
    {\footnotesize\begin{tablenotes}
        \item  The results are presented in terms of the average and the standard deviation (in bracket) of 100 replications.
    \end{tablenotes}}
\end{table}

\subsection{Real data example}

A real data set called Adult Data is considered, which is available in the UCI ML Repository, with the link \url{https://archive.ics.uci.edu/dataset/2/adult}\cite{misc_adult_2}. The data is extracted from the 1994 Census database. The main purpose of the data analysis is to predict the income. The variable used as the response in this data is the indicator indicating whether a person's annual income is larger than $\$50,000$, coded as $1$ for yes and $0$ for not. There are $48,842$ samples in all, with $11,687$ cases and $37,155$ controls. The variables on demographic information can be used as covariates. In the data there are $6$ continuous variables and $8$ categorical variables. We drop the three variables containing missing values. The variables ``relationship" and ``education" are excluded due to their high extent of multicollinearity. We also leave out the variables ``marial status" and ``fnlwgt" since they are insignificant for the response. The four categories within the ``race" variable referring to non-white people are consolidated into one category called ``non-white". Finally 7 covariates are included in the regression model.

We randomly select 10\% of the entire data set to be the testing set. For the remaining 90\%, we first draw a case-control sample to be the primary training set. Again, three sample sizes of case and control are considered. In the first one, $n_1=n_0=1000$. In the second one, $n_1=500$ and $n_0=1500$. In the third one, $n_1=200$ and $n_0=1800$. Then a random sample of size $2000$ is drawn to serve as the external data. We use the mean of the continuous covariate ``age" to be the external summary information. Meanwhile, we draw an independent case-control sample with the same size as the primary training set to be the validation set. The hyperparameters in the MLP are tuned based on this validation set in the same pattern to that in the simulation study. The learning rate is set to be 0.01 and the training process iterates for a maximum of 10,000 epochs. We obtain the proposed estimator $\widehat{g}_n(x)$ and the estimator without using external information $\widetilde{g}_n(x)$ based on the training set. The benchmark of the estimators is set to be the function trained based on the entire data. Specifically, leaving out the testing set, we use $80\%$ of the entire data set to be the training set and $10\%$ to be the validation set. The MLP is applied for functional approximation and the resultant neural network estimator, denoted by $\bar{g}(x)$, is treated as the benchmark. Under the three sample sizes, we obtain the scatter plots of $\widehat{g}_n(x)$ against $\bar{g}(x)$ and $\widetilde{g}_n(x)$ against $\bar{g}(x)$ evaluated on the testing set. The scatter plots are given in Figure \ref{fig2}.

From the plots, we can see that the scatter points of $\widehat{g}_n(x)$ and $\bar{g}(x)$ are primarily located near the 45-degree line under all the sample sizes. It means that the proposed estimator gives out quite close perdition results to that of the estimator based on the entire data. By contrast, the scatter points of $\widetilde{g}_n(x)$ and $\bar{g}(x)$ deviate from the 45-degree line, especially under the first and third sample size. Thus, using the case-control sample without the external information may bring bias in function estimation and probability prediction.

\begin{figure}[htbp]
    \centering
    \includegraphics[width=\textwidth]{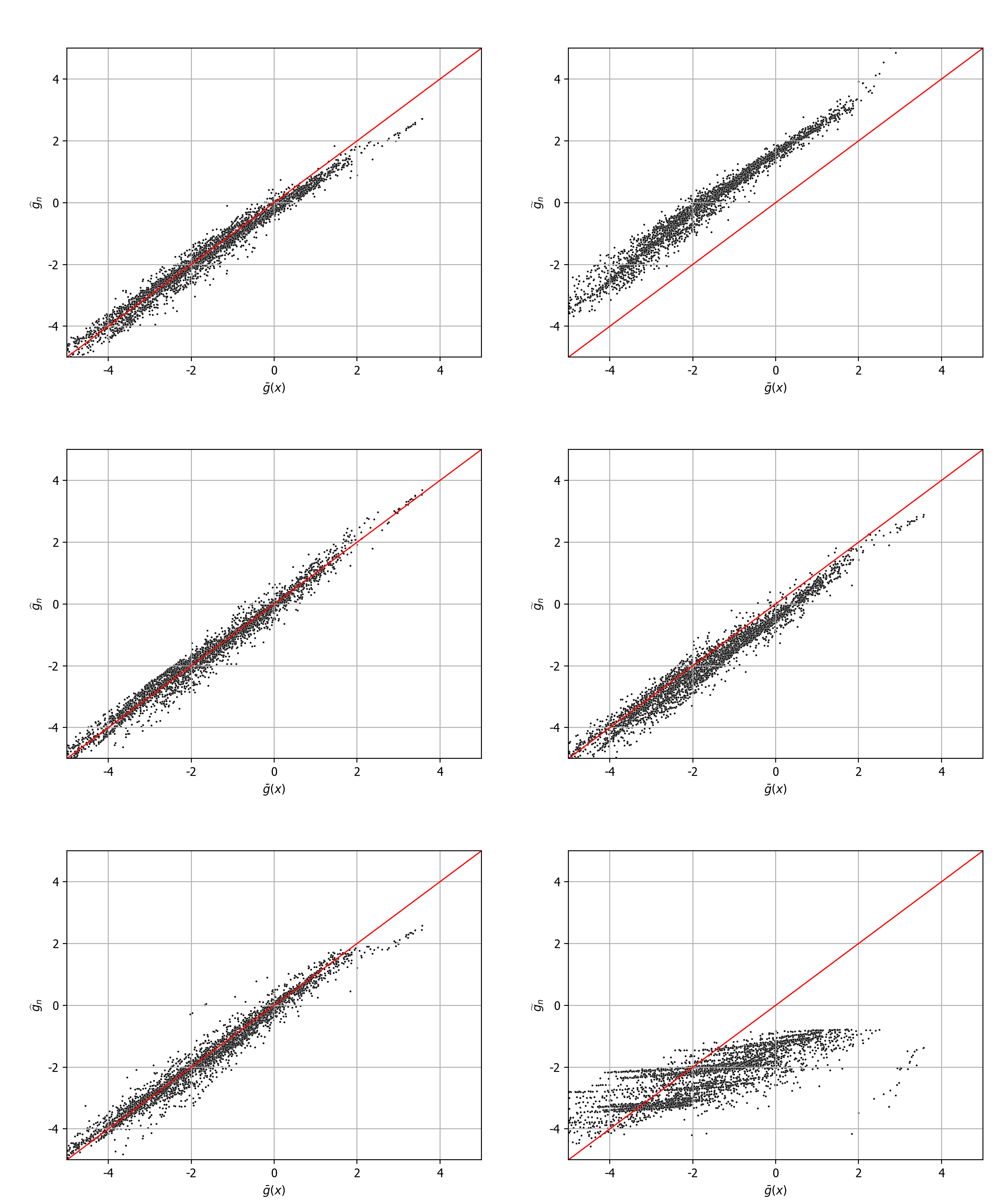}
    \caption{The scatter plots of the case-control data estimator against full data estimator. The left panel is for $\widehat{g}_n(x)$ and the right panel is for $\widetilde{g}_n(x)$. The three rows correspond to three case and control sizes.}
    \label{fig2}
    % \begin{tablenotes}
    %     \item  Note: 
    % \end{tablenotes}
\end{figure}

\section{Concluding remarks}\label{conclu}

We consider non-parametric estimation for the case-control logistic regression model. By utilizing external summary information, the model is identifiable. A two-step estimation procedure is developed. We first estimate the marginal case proportion by using external summary information. Then a weighted objective function is constructed based on the inverse probability weighting approach. The MLP is adopted to approximate the unknown function in the model. The deep structure provides the flexibility of the functional approximation. Some sophisticated theoretical results are established, including the non-asymptotic error bound of the excess risk and the convergence rate of the proposed estimator. The rate is shown to attain the optimal speed of the classical non-parametric regression estimation. The estimating procedure is quite easy to implement.

The non-parametric logistic model considered is quite general in the sense that it almost requires no specific structure on the conditional case probability. There are quite a lot of techniques for non-parametric estimation beside the neural networks. For instance, \cite{wang2023semi} adopted B-splines. Some comparison among different popular non-parametric methods is quite interesting. Meanwhile, the main idea of the proposed approach can be extended to other biased sampling scenario. All these guarantee some related future research.

\appendix

\section{Appendix}

%\subsection{Implementation details of computation}\label{a1}

\subsection{Proof of theorems and corollary}\label{a2}

\noindent{\it Proof of Theorem \ref{thm1}:} First we prove the consistency of $\hat{P}_1$. From the law of large number (LLN) we have that for $h(X) \in \mathbb{R}$ satisfying suitable regularity conditions, $\hat{\mu}_1 \xrightarrow{p} \mu_1$ and $\hat{\mu}_0 \xrightarrow{p} \mu_0$. Under Assumption \ref{ass4} and Assumption \ref{ass5}, by the continuous mapping theorem, we have that $\hat{P}_1 \xrightarrow{p} P_1$.

Next we show the asymptotic normality of $\hat{P}_1$. Define $H(x,y,z) = (x-z)/(y-z)$. Then $P_1$ and $\hat{P}_1$ can be written as $H(\mu,\mu_1,\mu_0)$ and $H(\tilde{\mu},\hat{\mu}_1,\hat{\mu}_0)$ respectively. According to the central limit theorem (CLT), we have that 
\begin{align*}
    \sqrt{n}(\hat{\mu}_1 - \mu_1) &\xrightarrow{d} N(0,V_1), \\
    \sqrt{n}(\hat{\mu}_0 - \mu_0) &\xrightarrow{d} N(0,V_0)
\end{align*}
as $n\to\infty$, where $V_1 = \lambda_1^{-1} \mathsf{Var}(h(X)|Y=1)$ and $V_0 = (1-\lambda_1)^{-1} \mathsf{Var}(h(X)|Y=0)$.

By Taylor's expansion, we have that
\begin{eqnarray}\label{f22}
\sqrt{n}H(\tilde{\mu},\hat{\mu}_1,\hat{\mu}_0)=\sqrt{n}H(\mu,\mu_1,\mu_0) + \sqrt{n}(a_1,a_2,a_3)(\hat{\mu}_1- \mu_1,\hat{\mu}_0 - \mu_0, \tilde{\mu} - \mu)^\top + o(1).
\end{eqnarray}
where $a_1,a_2,a_3$ ate the partial derivative of the $H(\mu,\mu_1,\mu_0)$ with respect to $\mu_1$, $\mu_0$, and $\mu$. After some detailed calculation, we can get 
\begin{eqnarray*}
    a_1 = \frac{\mu_0 - \mu}{(\mu_1 - \mu_0)^2}, a_2 = \frac{\mu - \mu_1}{(\mu_1 - \mu_0)^2}, a_3 = \frac{1}{\mu_1 - \mu_0}.
\end{eqnarray*}
Hence, using the independence among the three samples and (\ref{f22}), we have that 
\begin{align*}
    \sqrt{n}(\hat{P}_1 - P_1) \xrightarrow{d} N(0,a_1^2 V_1 + a_2^2 V_0 + a_3^2 V).
\end{align*}\qed

%在证明定理1之前，我们先给出一些引理来辅助我们的证明。我们有

Before proving Theorem \ref{thm2}, we first give out some lemmas to assist our proof. It is not difficult to show that 
\begin{align*}
    l(Z;g)&=\hat{w}_1^{-1}Y\log\psi(g(X))+\hat{w}_0^{-1}(1-Y)\log(1-\psi(g(X)))\\
        &= - \hat{w}_1^{-1} Y g(X) + [ \hat{w}_1^{-1} Y + \hat{w}_0^{-1} (1-Y) ] \log (1+e^{g(X)}).
\end{align*}

% lm1 ==========================================
\begin{lem}\label{lm1}
For training sample $\mathcal{S}$, we have 
    $$\mathsf{E}_\mathcal{S} \mathsf{E} r(Z;\hat{g}_n) \leq \mathsf{E}_\mathcal{S} [ \mathsf{E} r(Z;\hat{g}_n) - 2 \mathsf{E}_n r(Z;\hat{g}_n) ] + 2 \inf_{g \in \mathcal{G}_n} \mathsf{E} r(Z;g).$$
\end{lem}

\noindent{\it Proof of Lemma \ref{lm1}:} By the definition that $\hat{g}_n = \arg\min_{g \in \mathcal{G}_n} \mathsf{E}_n l(Z;g) $, $\mathsf{E} l(Z;\tilde{g}_n) = \inf_{g \in \mathcal{G}_n} \mathsf{E} r(Z;g)$. Since $\mathsf{E}_\mathcal{S} \mathsf{E}_n r(Z;\hat{g}_n) \leq \mathsf{E}_\mathcal{S} \mathsf{E}_n r(Z;\tilde{g}_n) = \mathsf{E} r(Z;\tilde{g}_n)$, we can get 
\begin{align*}
    \mathsf{E}_\mathcal{S} \mathsf{E} r(Z;\hat{g}_n) &= \mathsf{E}_\mathcal{S} [ \mathsf{E} r(Z;\hat{g}_n) - 2 \mathsf{E}_n r(Z;\hat{g}_n) ] + 2 \mathsf{E}_\mathcal{S} \mathsf{E}_n r(Z;\hat{g}_n) \\
    &\leq \mathsf{E}_\mathcal{S} [ \mathsf{E} r(Z;\hat{g}_n) - 2 \mathsf{E}_n r(Z;\hat{g}_n) ] + 2\mathsf{E} r(Z;\tilde{g}_n)\\
    % &= \mathsf{E}_\mathcal{S} [ \mathsf{E}_Z r(Z;\hat{g}_n) - 2 \mathsf{E}_n r(Z;\hat{g}_n) ] + 2 \inf_{g \in \mathcal{G}_n} \mathsf{E}_Z r(Z;g)\\
    &= \underbrace{\mathsf{E}_\mathcal{S} [ \mathsf{E} r(Z;\hat{g}_n) - 2 \mathsf{E}_n r(Z;\hat{g}_n) ]}_{\text{stochastic error}} + \underbrace{2 \inf_{g \in \mathcal{G}_n} \mathsf{E} r(Z;g)}_{\text{approximation error}}. 
\end{align*}\qed

% lm2 ==========================================2\hat{w}_1^{-1} +\hat{w}_0^{-1}
\begin{lem}\label{lm2}
    Suppose that $\mathcal{G}_n$ consists of MLPs with width $W_n = 38\beta^2N_n\lceil\log_2(8N_n)\rceil$ and depth $D_n = 21\beta^2M_n\lceil\log_2(8M_n)\rceil$ for any $M_n,N_n \in \mathbb{N}^+$.Then 
    \begin{align}\label{f16}
        \inf_{g \in \mathcal{G}_n} \mathsf{E} r(Z;g) \leq 18(2\hat{w}_1^{-1} +\hat{w}_0^{-1})B\beta^2 p^{3\beta/2}(NM)^{-2\beta/p}.
    \end{align}
\end{lem}

\noindent{\it Proof:}
By the definition $r(Z;g) = l(Z;g) - l(Z;g_0)$, we have
\begin{align}\label{f15}
    \mathsf{E} r(Z;g) &= \mathsf{E} \left[-\hat{w}_1^{-1} Y (g(X) - g_0(X)) + [ \hat{w}_1^{-1} Y + \hat{w}_0^{-1} (1-Y) ] [\log (1+e^{g(X)}) - \log (1+e^{g_0(X)})] \right] \notag\\
        &\leq \mathsf{E} \left[ \hat{w}_1^{-1}|g(X) - g_0(X)| + (\hat{w}_1^{-1} + \hat{w}_0^{-1})|\log (1+e^{g(X)}) - \log (1+e^{g_0(X)})|\right] \notag \\
        &\leq \mathsf{E} \left[ \hat{w}_1^{-1}|g(X) - g_0(X)| + (\hat{w}_1^{-1} + \hat{w}_0^{-1}) |g(X) - g_0(X)| \right] \notag \\
        &= \mathsf{E}_X \left[ (2\hat{w}_1^{-1} + \hat{w}_0^{-1})|g(X) - g_0(X)|\right],
\end{align}
where the second inequality comes from $| \log (1+e^a) - \log (1+e^b) | \leq |a - b|$ for any $a,b$. 
By Theorem 3.3 of \cite{jiao2023deep}, there exists a function $g^* \in \mathcal{G}_n$ such that 
\begin{align}\label{f14}
    |g^*(x)-g_0(x)| \leq 18B\beta^2 p^{3/2}(N_n M_n)^{-2\beta/p}
\end{align}
for all $x \in [0,1]^d\backslash \Omega([0,1]^d,K,\delta)$, where 
\begin{align*}
    \Omega([0,1]^d,K,\delta) = \bigcup_{i=1}^d {x = [x_1,\dots,x_d]^T : x_i \in \bigcup_{k=1}^K (k/K - \delta, k/K)}
\end{align*}
with $K = \lceil (N_n M_n)^{2/d} \rceil$ and $\delta$ an arbitrary number in $(0,1/(3K)]$.\\
Combine (\ref{f15}) and (\ref{f14}), we have
\begin{align*}
    \mathsf{E}_X \left[ (2\hat{w}_1^{-1} + \hat{w}_0^{-1})|g^*(X) - g_0(X)|\right] \leq (2\hat{w}_1^{-1} + \hat{w}_0^{-1})[18B\beta^2 p^{3/2}(N_n M_n)^{-2\beta/p} + 2c_1 B \mu(\Omega([0,1]^d,K,\delta)) ].
\end{align*}
where $\mu(\Omega([0,1]^d,K,\delta))$ is the Lebesgue measure of $\Omega([0,1]^d,K,\delta)$, which is no more than $dK\delta$. Similar to the proof of Theorem 4.2 in \citep{jiao2023deep}, $\delta$ can be sufficiently small and under Assumption \ref{ass1}, we give out that
\begin{align*}
    \inf_{g \in \mathcal{G}_n} \mathsf{E} r(Z;g) \leq 18(2\hat{w}_1^{-1} + \hat{w}_0^{-1})B\beta^2 p^{3/2}(N_n M_n)^{-2\beta/p}.
\end{align*}\qed

% lm3 ==========================================
\begin{lem}\label{lm3}
    $|r(Z;g)| \leq 2B(2\hat{w}_1^{-1} + \hat{w}_0^{-1}) $.
\end{lem}

\noindent{\it Proof:} 
\begin{align*}
    r(Z;g) &= -\hat{w}_1^{-1} Y (g(X) - g_0(X)) + [ \hat{w}_1^{-1} Y + \hat{w}_0^{-1} (1-Y) ] [\log (1+e^{g(X)}) - \log (1+e^{g_0(X)})] \\
        &\leq \hat{w}_1^{-1}|g(X) - g_0(X)| + (\hat{w}_1^{-1} + \hat{w}_0^{-1}) |g(X) - g_0(X)| \\
        &\leq 2B(2\hat{w}_1^{-1} + \hat{w}_0^{-1}),
\end{align*}
where the first inequality comes from $| \log (1+e^a) - \log (1+e^b) | \leq |a - b|$ for any $a,b$ and $Y \in \{0,1\}$.\qed

% lm ==========================================
\begin{lem}\label{lm4}
    $\mathsf{Var}(r(Z;g)) \leq \mathsf{E}(r(Z;g))^2 \leq 2B(2\hat{w}_1^{-1} + \hat{w}_0^{-1})\mathsf{E}(r(Z;g))$.
\end{lem}

\noindent{\it Proof:} Obviously the first inequality holds true. By Lemma \ref{lm3} we have $r(Z;g)^2 \leq 2B(2\hat{w}_1^{-1} + \hat{w}_0^{-1})r(Z;g)$, hence $\mathsf{E}(r(Z;g))^2 \leq 2B(2\hat{w}_1^{-1} + \hat{w}_0^{-1})\mathsf{E}(r(Z;g))$.\qed

Next we give an upper bound of the covering number of $\mathcal{R}_n = \{r(Z;g): g \in \mathcal{G}_n\}$. 

% lm ==========================================
\begin{lem}\label{lm5}
    For the covering number about $\mathcal{G}_n$ and $\mathcal{R}_n$, we have
    \begin{align*}
        \mathcal{N}_1 (1/n , \mathcal{R}_n,Z_1^n) \leq \mathcal{N}_1 (1/(2\hat{w}_1^{-1} + \hat{w}_0^{-1})n , \mathcal{G}_n,Z_1^n).
    \end{align*}
\end{lem}

\noindent{\it Proof:} Let $g_1,\dots,g_l, l = \mathcal{N}_1 (1/(2\hat{w}_1^{-1} + \hat{w}_0^{-1})n , \mathcal{G}_n,Z_1^n)$ be an  $1/(2\hat{w}_1^{-1} + \hat{w}_0^{-1})n$-cover of $\mathcal{G}_n$ on $Z_1^n$. Let $g \in \mathcal{G}_n$ be arbitrary. Then there exists an $g_j$ such that $\frac{1}{n} \sum_{i=1}^n |g(X_i) - g_j (X_i)| < 1/{(2\hat{w}_1^{-1} + \hat{w}_0^{-1})n}$. We have 
\begin{align*}
    \frac{1}{n} \sum_{i=1}^n |r(Z_i ; g) - r(Z_i ; g_j)| &= \frac{1}{n} \sum_{i=1}^n |l(Z_i ; g) - l(Z_i ; g_0) - l(Z_i ; g_j)+l(Z_i ; g_0)| \\
                        &=\frac{1}{n} \sum_{i=1}^n |-\hat{w}_1^{-1} Y_i (g - g_j) + [\hat{w}_1^{-1} Y_i +\hat{w}_0^{-1} (1-Y_i)](\log(1+e^{g})-\log(1+e^{g_j}))| \\
                        &\leq\frac{1}{n} \sum_{i=1}^n (2\hat{w}_1^{-1} + \hat{w}_0^{-1})|g(X_i) - g_j(X_i)| \\
                        &\leq \frac{1}{n}.
\end{align*}
Thus $g_1,\dots,g_l$ is an $\frac{1}{n}-$cover of $\mathcal{R}_n$ on $Z_1^n$ of size $\mathcal{N}_1 (1/(2\hat{w}_1^{-1} + \hat{w}_0^{-1})n , \mathcal{G}_n,Z_1^n)$.\qed

% lm ==========================================
\begin{lem}\label{lm6}
Assume $B\geq 5/4(2\hat{w}_1^{-1} + \hat{w}_0^{-1})$, for $\forall \alpha,\beta >0$ and $0<\epsilon<1/2$,
    \begin{align}\label{a}
        &\mathsf{P}  \left( \exists g \in \mathcal{G}_n : \mathsf{E}r(Z;g) - \frac{1}{n} \sum_{i=1}^{n} r(Z_i;g) \geq \epsilon(\alpha + \beta + \mathsf{E}r(Z;g)) \right) \notag \\
        &\qquad \leq 14 \mathcal{N}_1 (\frac{\epsilon \beta}{5},\mathcal{R}_n,Z_1^n) \exp \left( -\frac{\epsilon^2 (1-\epsilon) \alpha n}{54 B^2(2\hat{w}_1^{-1} +\hat{w}_0^{-1})^2 (1+\epsilon)}\right).
    \end{align}
\end{lem}

\noindent{\it Proof:} We adopt the same technique used in the proof of Theorem 11.4 of \cite{gyorfi2002distribution} and extend it beyond the non-parametric regression.

First, replace the expectation on the left-hand side of the inequality in (\ref{a}) by the empirical mean based on $\{ Z_i^\prime = (X_i^\prime,Y_i^\prime)\}_{i=1}^{n}$ of another i.i.d copies of $Z$ and is independent of $Z_1^n$. Consider a function $g_n \in \mathcal{G}_n$ depending upon $Z_1^n$ such that 
\begin{align*}
    \mathsf{E}[r(Z;g_n)|Z_1^n] - \frac{1}{n} \sum_{i=1}^{n} r(Z_i;g_n) \geq \epsilon(\alpha + \beta) + \epsilon \mathsf{E}[r(Z;g_n)|Z_1^n]
\end{align*}
if such a function exists in $\mathcal{G}_n$, otherwise choose an arbitrary function in $\mathcal{G}_n$. By Lemma \ref{lm4} and Chebyshev's inequality, we have 
\begin{align*}
    &\mathsf{P} \left(  \mathsf{E}[r(Z;g_n)|Z_1^n] - \frac{1}{n} \sum_{i=1}^{n} r(Z_i^\prime;g_n) \geq \frac{\epsilon}{2} (\alpha + \beta) + \frac{\epsilon}{2} \mathsf{E}[r(Z;g_n)|Z_1^n]  \bigg| Z_1^n   \right) \\
    &\leq \frac{\mathsf{Var(r(Z;g_n))}|Z_1^n}{n \left( \frac{\epsilon}{2} (\alpha + \beta) + \frac{\epsilon}{2} \mathsf{E}[r(Z;g_n)|Z_1^n] \right)^2} \\
    &\leq \frac{2B(2 \hat{w}_1^{-1} + \hat{w}_0^{-1}) \mathsf{E}[r(Z;g_n)|Z_1^n]}{n \left( \frac{\epsilon}{2} (\alpha + \beta) + \frac{\epsilon}{2} \mathsf{E}[r(Z;g_n)|Z_1^n] \right)^2} \\
    &\leq \frac{2B(2 \hat{w}_1^{-1} + \hat{w}_0^{-1})}{n\epsilon^2 (\alpha + \beta)},
\end{align*}
where the last inequality follows from $x/(a+x)^2 \leq 1/4a $ for all $x,a > 0$. Thus for $n > {16B(2 \hat{w}_1^{-1} + \hat{w}_0^{-1})}/{\epsilon^2 (\alpha + \beta)} $, we have 
\begin{align}\label{b}
    \mathsf{P} \left(  \mathsf{E}[r(Z;g_n)|Z_1^n] - \frac{1}{n} \sum_{i=1}^{n} r(Z_i^\prime;g_n) \leq \frac{\epsilon}{2} (\alpha + \beta) + \frac{\epsilon}{2} \mathsf{E}[r(Z;g_n)|Z_1^n]  \bigg| Z_1^n   \right) \geq \frac{7}{8}.
\end{align}
Hence,
\begin{align*}
    &\mathsf{P} \left(  \exists g \in \mathcal{G}_n : \frac{1}{n} \sum_{i=1}^{n} r(Z_i^\prime;g) - \frac{1}{n} \sum_{i=1}^{n} r(Z_i;g) \geq \frac{\epsilon}{2} (\alpha + \beta) + \frac{\epsilon}{2} \mathsf{E}r(Z;g)  \right) \\
    &\geq \mathsf{P} \left( \frac{1}{n} \sum_{i=1}^{n} r(Z_i^\prime;g_n) - \frac{1}{n} \sum_{i=1}^{n} r(Z_i;g_n) \geq \frac{\epsilon}{2} (\alpha + \beta) + \frac{\epsilon}{2} \mathsf{E}[r(Z;g_n)|Z_1^n]  \right) \\
    &\geq \mathsf{P} \left( \mathsf{E}[r(Z;g_n)|Z_1^n] - \frac{1}{n} \sum_{i=1}^{n} r(Z_i;g_n) \geq \epsilon(\alpha + \beta ) + \epsilon \mathsf{E}[r(Z;g_n)|Z_1^n], \right.\\
    &\quad\quad\quad \left. \mathsf{E}[r(Z;g_n)|Z_1^n] -  \sum_{i=1}^{n} r(Z_i^\prime;g_n) \leq  \frac{\epsilon}{2} (\alpha + \beta) + \frac{\epsilon}{2} \mathsf{E}[r(Z;g_n)|Z_1^n] \right) \\
    &= \mathsf{E} \left[ I_{\{\mathsf{E}[r(Z;g_n)|Z_1^n] - \frac{1}{n} \sum_{i=1}^{n} r(Z_i;g_n) \geq \epsilon(\alpha + \beta ) + \epsilon \mathsf{E}[r(Z;g_n)|Z_1^n]\}} \times \right.\\
    &\quad\quad\quad \left.  \mathsf{P} \left( \mathsf{E}[r(Z;g_n)|Z_1^n] -  \sum_{i=1}^{n} r(Z_i^\prime;g_n) \leq  \frac{\epsilon}{2} (\alpha + \beta) + \frac{\epsilon}{2} \mathsf{E}[r(Z;g_n)|Z_1^n]\right) \right] \\
    &\geq \frac{7}{8} \mathsf{P}  \left(  \mathsf{E}[r(Z;g_n)|Z_1^n] - \frac{1}{n} \sum_{i=1}^{n} r(Z_i;g_n) \geq \epsilon(\alpha + \beta ) + \epsilon \mathsf{E}[r(Z;g_n)|Z_1^n] \right) \\
    &= \frac{7}{8} \mathsf{P} \left(  \exists g \in \mathcal{G}_n : \mathsf{E}[r(Z;g)|Z_1^n] - \frac{1}{n} \sum_{i=1}^{n} r(Z_i;g) \geq \epsilon(\alpha + \beta ) + \epsilon \mathsf{E}[r(Z;g)|Z_1^n] \right).
\end{align*}
Thus, for $n > {16B(2\hat{w}_1^{-1} + \hat{w}_0^{-1})}/{\epsilon^2 (\alpha + \beta)}$,
\begin{align}\label{c}
    &\mathsf{P} \left(  \exists g \in \mathcal{G}_n : \mathsf{E}[r(Z;g)|Z_1^n] - \frac{1}{n} \sum_{i=1}^{n} r(Z_i;g) \geq \epsilon(\alpha + \beta ) + \epsilon \mathsf{E}[r(Z;g)|Z_1^n] \right) \notag \\ 
    &\leq \frac{8}{7} \mathsf{P} \left(  \exists g \in \mathcal{G}_n : \frac{1}{n} \sum_{i=1}^{n} r(Z_i^\prime;g) - \frac{1}{n} \sum_{i=1}^{n} r(Z_i;g) \geq \frac{\epsilon}{2} (\alpha + \beta) + \frac{\epsilon}{2} \mathsf{E}r(Z;g)  \right).
\end{align}
Secondly, we control the right-hand side of (\ref{c}),
\begin{align}\label{d}
    &\mathsf{P} \left(  \exists g \in \mathcal{G}_n : \frac{1}{n} \sum_{i=1}^{n} r(Z_i^\prime;g) - \frac{1}{n} \sum_{i=1}^{n} r(Z_i;g) \geq \frac{\epsilon}{2} (\alpha + \beta) + \frac{\epsilon}{2} \mathsf{E}r(Z;g)  \right) \notag \\
    &\leq \mathsf{P} \left(  \exists g \in \mathcal{G}_n : \frac{1}{n} \sum_{i=1}^{n} r(Z_i^\prime;g) - \frac{1}{n} \sum_{i=1}^{n} r(Z_i;g) \geq \frac{\epsilon}{2} (\alpha + \beta) + \frac{\epsilon}{2} \mathsf{E}r(Z;g), \right. \notag \\
    &\hspace{2cm} \left. \frac{1}{n} \sum_{i=1}^{n} r^2(Z_i;g) - \mathsf{E} r^2(Z;g) \leq \epsilon (\alpha + \beta + \frac{1}{n} \sum_{i=1}^{n} r^2(Z_i;g) + \mathsf{E} r^2(Z;g)), \right. \notag \\
    &\hspace{2cm} \left. \frac{1}{n} \sum_{i=1}^{n} r^2(Z_i^\prime;g) - \mathsf{E} r^2(Z;g) \leq \epsilon (\alpha + \beta + \frac{1}{n} \sum_{i=1}^{n} r^2(Z_i^\prime;g) + \mathsf{E} r^2(Z;g)) \right) \notag \\
    &\quad + 2 \mathsf{P} \left( \exists g \in \mathcal{G}_n : \frac{\frac{1}{n} \sum_{i=1}^{n} r^2(Z_i;g) - \mathsf{E} r^2(Z;g)}{\alpha + \beta + \frac{1}{n} \sum_{i=1}^{n} r^2(Z_i;g) + \mathsf{E} r^2(Z;g)} > \epsilon \right).
\end{align}
By Lemma \ref{lm3} and Theorem 11.6 of \cite{gyorfi2002distribution}, the second probability on the right-hand side of (\ref{d}) yields
\begin{align}\label{e}
    &\mathsf{P} \left( \exists g \in \mathcal{G}_n : \frac{\frac{1}{n} \sum_{i=1}^{n} r^2(Z_i;g) - \mathsf{E} r^2(Z;g)}{\alpha + \beta + \frac{1}{n} \sum_{i=1}^{n} r^2(Z_i;g) + \mathsf{E} r^2(Z;g)} > \epsilon \right) \notag\\
    &\leq 4 \mathcal{N}_1 (\frac{\epsilon (\alpha + \beta)}{5}, \mathcal{R}_n, Z_1^n) \exp \left( -\frac{3 \epsilon^2 (\alpha + \beta )n}{160B^2(2\hat{w}_1^{-1} + \hat{w}_0^{-1})^2}\right).
\end{align}
Now we consider the first probability on the right-hand side of (\ref{d}), the second inequality inside the probability implies 
\begin{align}\label{f}
    (1 + \epsilon) \mathsf{E} r^2(Z;g) \geq (1- \epsilon) \frac{1}{n} \sum_{i=1}^{n} r^2(Z_i;g) - \epsilon (\alpha + \beta).
\end{align}
We can deal similarly with the third inequality. Using (\ref{f}) and Lemma \ref{lm4} we can bound the first probability on the right-hand side of (\ref{d}) by 
\begin{align}\label{i}
    &\mathsf{P} \left(  \exists g \in \mathcal{G}_n : \frac{1}{n} \sum_{i=1}^{n} r(Z_i^\prime;g) - \frac{1}{n} \sum_{i=1}^{n} r(Z_i;g) \geq \frac{\epsilon}{2} (\alpha + \beta) \right. \notag\\
    &\hspace{4cm} \left. + \frac{\epsilon}{2} \left[  \frac{1-\epsilon}{4B(2\hat{w}_1^{-1} + \hat{w}_0^{-1})(1 + \epsilon)}\frac{1}{n} \sum_{i=1}^{n} r^2(Z_i;g) - \frac{\epsilon (\alpha + \beta)}{4B(2\hat{w}_1^{-1} + \hat{w}_0^{-1})(1 + \epsilon)}   \right.\right. \notag \\
    &\hspace{5cm} \left.\left. + \frac{1-\epsilon}{4B(2\hat{w}_1^{-1} + \hat{w}_0^{-1})(1 + \epsilon)}\frac{1}{n} \sum_{i=1}^{n} r^2(Z_i^\prime;g) - \frac{\epsilon (\alpha + \beta)}{4B(2\hat{w}_1^{-1} + \hat{w}_0^{-1})(1 + \epsilon)}  \right]\right).
\end{align}
Let $U_1,\dots,U_n$ be independent and uniformly distributed over the set $\{-1,1\}$ and independent of $Z_1,\dots,Z_n , Z_1^\prime , \dots , Z_n^\prime$. So $Z_1^{\prime n}$ is not affected by the random interchange of the corresponding components in $Z_1$ and $Z_n$. Therefore (\ref{i}) is equal to 
\begin{align}\label{j}
    &\mathsf{P} \left(  \exists g \in \mathcal{G}_n : \frac{1}{n} \sum_{i=1}^{n} U_i ( r(Z_i^\prime;g) - r(Z_i;g) )\geq  \frac{\epsilon}{2} (\alpha + \beta) -\frac{\epsilon^2 (\alpha + \beta)}{4B(2\hat{w}_1^{-1} + \hat{w}_0^{-1})(1 + \epsilon)} \right. \notag\\
    &\hspace{5cm} \left.+  \frac{\epsilon(1-\epsilon)}{8B(2\hat{w}_1^{-1} + \hat{w}_0^{-1})(1 + \epsilon)}\frac{1}{n} \sum_{i=1}^{n} [r^2(Z_i^\prime;g) + r^2(Z_i;g) ] \right).
\end{align}
and this (\ref{j}), in turn, by the union bound, is bounded by 
\begin{align}\label{k}
    &\mathsf{P} \left(  \exists g \in \mathcal{G}_n : \left| \frac{1}{n}  \sum_{i=1}^{n} U_i r(Z_i^\prime;g) \right| \geq \frac{1}{2}\left[ \frac{\epsilon}{2} (\alpha + \beta) - \frac{\epsilon^2 (\alpha + \beta)}{4B(2\hat{w}_1^{-1} + \hat{w}_0^{-1})(1 + \epsilon)} \right] \right. \notag \\ 
    &\hspace{6cm} \left. + \frac{\epsilon(1-\epsilon)}{8B(2\hat{w}_1^{-1} + \hat{w}_0^{-1})(1 + \epsilon)}\frac{1}{n} \sum_{i=1}^{n} r^2(Z_i^\prime;g) \right) \notag \\
    &+ \mathsf{P} \left(  \exists g \in \mathcal{G}_n : \left| \frac{1}{n}  \sum_{i=1}^{n} U_i r(Z_i;g) \right| \geq \frac{1}{2}\left[ \frac{\epsilon}{2} (\alpha + \beta) - \frac{\epsilon^2 (\alpha + \beta)}{4B(2\hat{w}_1^{-1} + \hat{w}_0^{-1})(1 + \epsilon)} \right] \right. \notag \\ 
    &\hspace{6cm} \left. +\frac{\epsilon(1-\epsilon)}{8B(2\hat{w}_1^{-1} + \hat{w}_0^{-1})(1 + \epsilon)}\frac{1}{n} \sum_{i=1}^{n} r^2(Z_i;g) \right) \notag \\
    &= 2 \mathsf{P} \left(  \exists g \in \mathcal{G}_n : \left| \frac{1}{n}  \sum_{i=1}^{n} U_i r(Z_i;g) \right| \geq  \frac{\epsilon}{4} (\alpha + \beta) - \frac{\epsilon^2 (\alpha + \beta)}{8B(2\hat{w}_1^{-1} + \hat{w}_0^{-1})(1 + \epsilon)}  \right. \notag \\ 
    &\hspace{6cm} \left. + \frac{\epsilon(1-\epsilon)}{8B(2\hat{w}_1^{-1} + \hat{w}_0^{-1})(1 + \epsilon)}\frac{1}{n} \sum_{i=1}^{n} r^2(Z_i;g) \right).
\end{align}
Let $\delta > 0$ and $\mathcal{R}_n^\delta$ be an $L_1 - \delta$ cover of $\mathcal{R}_n = \{r(Z;g): g \in \mathcal{G}_n\}$ on $Z_1^n$, which is equivalent to fixing $Z_i$, and denote the corresponding $\mathcal{G}_n^\delta = \{g : r(Z;g) \in \mathcal{R}_n^\delta\}$. For $g \in \mathcal{G}_n$, there exists $g^\prime \in \mathcal{G}_n^\delta$ such that 
\begin{align*}
    \frac{1}{n} \sum_{i=1}^{n} |r(Z_i;g) - r(Z_i;g^\prime)| < \delta.
\end{align*}
This implies
\begin{align}\label{g}
    \left|\frac{1}{n} \sum_{i=1}^{n} U_i r(Z_i;g)\right| &= \left|\frac{1}{n} \sum_{i=1}^{n} U_i r(Z_i;g^\prime) + \frac{1}{n} \sum_{i=1}^{n} U_i ( r(Z_i;g) - r(Z_i;g^\prime) )\right| \notag \\
    &\leq \left|\frac{1}{n} \sum_{i=1}^{n} U_i r(Z_i;g^\prime)\right| + \delta
\end{align}
and 
\begin{align}\label{h}
    \left| \frac{1}{n} \sum_{i=1}^{n} r^2(Z_i;g) \right| &= \left| \frac{1}{n} \sum_{i=1}^{n} r^2(Z_i;g^\prime) + \frac{1}{n} \sum_{i=1}^{n} (r^2(Z_i;g) - r^2(Z_i;g^\prime)) \right|  \notag \\
    &\geq \left| \frac{1}{n} \sum_{i=1}^{n} r^2(Z_i;g^\prime) \right| -4B(2\hat{w}_1^{-1} + \hat{w}_0^{-1})\delta.
\end{align}
It follows the right-hand side of (\ref{k}) that 
\begin{align}\label{l}
    &\mathsf{P} \left(  \exists g \in \mathcal{G}_n : \left| \frac{1}{n} \sum_{i=1}^{n} U_i r(Z_i;g) \right| \geq  \frac{\epsilon}{4} (\alpha + \beta) - \frac{\epsilon^2 (\alpha + \beta)}{8B(2\hat{w}_1^{-1} + \hat{w}_0^{-1})(1 + \epsilon)}  + \frac{\epsilon(1-\epsilon)}{8B(2\hat{w}_1^{-1} + \hat{w}_0^{-1})(1 + \epsilon)}\frac{1}{n} \sum_{i=1}^{n} r^2(Z_i;g) \right) \notag \\
    &\leq \mathsf{P} \left(  \exists g \in \mathcal{G}_n^\delta : \left| \frac{1}{n} \sum_{i=1}^{n} U_i r(Z_i;g) \right| + \delta \geq  \frac{\epsilon}{4} (\alpha + \beta) - \frac{\epsilon^2 (\alpha + \beta)}{8B(2\hat{w}_1^{-1} + \hat{w}_0^{-1})(1 + \epsilon)}  \right. \notag \\
    &\hspace{6cm} \left. + \frac{\epsilon(1-\epsilon)}{8B(2\hat{w}_1^{-1} + \hat{w}_0^{-1})(1 + \epsilon)} \left[ \frac{1}{n} \sum_{i=1}^{n} r^2(Z_i;g) - 4B(2\hat{w}_1^{-1} + \hat{w}_0^{-1})\delta \right]\right) \notag \\
    &\leq |\mathcal{G}_n^\delta| \max_{g \in \mathcal{G}_n^\delta} \mathsf{P} \left( \left| \frac{1}{n} \sum_{i=1}^{n} U_i r(Z_i;g) \right| \geq \frac{\epsilon}{4} (\alpha + \beta) - \frac{\epsilon^2 (\alpha + \beta)}{8B(2\hat{w}_1^{-1} + \hat{w}_0^{-1})(1 + \epsilon)} - \delta - \frac{ \epsilon (1-\epsilon)}{2(1+ \epsilon)} \delta \right. \notag \\
    &\hspace{6cm} \left. + \frac{\epsilon(1-\epsilon)}{8B(2\hat{w}_1^{-1} + \hat{w}_0^{-1})(1 + \epsilon)}\frac{1}{n} \sum_{i=1}^{n} r^2(Z_i;g) \right).
\end{align}
Then we set $\delta = \epsilon \beta / 5$ and $B \geq 5/4 (2\hat{w}_1^{-1} + \hat{w}_0^{-1} )$, together with $0< \epsilon < 1/2$, we have that 
\begin{align*}
    \frac{\epsilon \beta}{5} - \frac{\epsilon^2 \beta}{8B(2\hat{w}_1^{-1} + \hat{w}_0^{-1})(1+ \epsilon)} - 
    \delta - \frac{\epsilon(1-\epsilon)}{2(1+\epsilon)}\delta \geq 0.
\end{align*}
Thus we consider to bound the probability on right-hand side of (\ref{l}) by 
\begin{align}\label{m}
    \mathsf{P} \left( \left| \frac{1}{n} \sum_{i=1}^{n} U_i r(Z_i;g) \right| \geq \frac{\epsilon \alpha}{4} - \frac{\epsilon^2 \alpha}{8B(2\hat{w}_1^{-1} + \hat{w}_0^{-1})(1 + \epsilon)} + \frac{\epsilon(1-\epsilon)}{8B(2\hat{w}_1^{-1} + \hat{w}_0^{-1})(1 + \epsilon)}\frac{1}{n} \sum_{i=1}^{n} r^2(Z_i;g)\right)
\end{align}
and we use Bernstein's inequality to bound (\ref{m}). First we relate $n^{-1} \sum_{i=1}^{n} r^2(Z_i;g)$ to the variance of $U_i r(Z_i;g)$,
\begin{align*}
    \frac{1}{n} \sum_{i=1}^{n} \mathsf{Var}(U_i r(Z_i;g)) = \frac{1}{n} \sum_{i=1}^{n} r^2(Z_i;g)\mathsf{Var}(U_i) = \frac{1}{n} \sum_{i=1}^{n} r^2(Z_i;g).
\end{align*}
Thus (\ref{m}) is equal to 
\begin{align*}%\label{o}
    \mathsf{P} \left( \left| \frac{1}{n} \sum_{i=1}^{n} V_i \right| \geq A_1 + A_2 \sigma^2 \right),
\end{align*}
where $V_i = U_i r(Z_i;g), \sigma^2= \frac{1}{n} \sum_{i=1}^{n} \mathsf{Var}(U_i r(Z_i;g)), A_1 = \frac{\epsilon \alpha}{4} - \frac{\epsilon^2 \alpha}{8B(2\hat{w}_1^{-1} + \hat{w}_0^{-1})(1 + \epsilon)}, A_2 = \frac{\epsilon(1-\epsilon)}{8B(2\hat{w}_1^{-1} + \hat{w}_0^{-1})(1 + \epsilon)}$
By Bernstein's inequality and $|V_i| \leq |r(Z_i;g)| \leq 2B(2\hat{w}_1^{-1} + \hat{w}_0^{-1})$, we have 
\begin{align}\label{p}
    &\mathsf{P} \left( \left| \frac{1}{n} \sum_{i=1}^{n} V_i \right| \geq A_1 + A_2 \sigma^2 \right) \notag \\
    &\leq 2\exp \left( -\frac{n{(A_1 + A_2 \sigma^2)}^2}{2\sigma^2 + \frac{4}{3}B(2\hat{w}_1^{-1} + \hat{w}_0^{-1})(A_1 + A_2 \sigma^2)}\right) \notag \\
    &= 2\exp \left( -\frac{3nA_2}{4B(2\hat{w}_1^{-1} + \hat{w}_0^{-1})} \cdot \frac{{(\frac{A_1}{A_2} + \sigma^2)}^2}{\frac{A_1}{A_2} + (1+\frac{3}{2B(2\hat{w}_1^{-1} + \hat{w}_0^{-1})A_2}\sigma^2)}\right).
\end{align}
Using the fact that for $\forall a,b,u > 0$,
\begin{align*}%\label{q}
    \frac{(a+u)^2}{a+bu} \geq 4a\frac{b-1}{b^2}.
\end{align*}
Set $a = A_1/A_2$, $b = 1+ 3/[2B(2\hat{w}_1^{-1} + \hat{w}_0^{-1})A_2]$, and $u = \sigma^2$. For the exponent in (\ref{p}), we obtain that
\begin{align*}%\label{r}
    -\frac{3nA_2}{4B(2\hat{w}_1^{-1} + \hat{w}_0^{-1})} \cdot \frac{{(\frac{A_1}{A_2} + \sigma^2)}^2}{\frac{A_1}{A_2} + (1+\frac{3}{2B(2\hat{w}_1^{-1} + \hat{w}_0^{-1})A_2}\sigma^2)} \geq \frac{9n A_1 A_2}{2(2B(2\hat{w}_1^{-1} + \hat{w}_0^{-1})A_2 + 3)^2}. 
\end{align*}
Substituting the formulas for $A_1$ and $A_2$ and noting that
\begin{align*}%\label{s}
    A_1 = \frac{\epsilon \alpha}{4} - \frac{\epsilon^2 \alpha}{8B(2\hat{w}_1^{-1} + \hat{w}_0^{-1})(1 + \epsilon)} \geq \frac{\epsilon \alpha}{4} - \frac{\epsilon^2 \alpha}{10(1+\epsilon)} \geq \frac{3\epsilon \alpha}{20},
\end{align*}
we obtain that
\begin{align*}%\label{t}
    \frac{9n A_1 A_2}{2(2B(2\hat{w}_1^{-1} + \hat{w}_0^{-1})A_2 + 3)^2} &\geq \frac{9n}{2} \cdot \frac{3\epsilon \alpha}{20} \cdot \frac{\epsilon(1-\epsilon)}{8B(2\hat{w}_1^{-1} + \hat{w}_0^{-1})(1+\epsilon)} \cdot \frac{1}{(\frac{\epsilon(1-\epsilon)}{4(1+\epsilon)}+3)^2}\\
                                    &\geq \frac{27n \epsilon^2 (1-\epsilon) \alpha}{320 B (2\hat{w}_1^{-1} + \hat{w}_0^{-1})(1+\epsilon) }\cdot \frac{1}{(\frac{1}{16}+3)^2} \\
                                    &= \frac{n \epsilon^2 (1-\epsilon) \alpha}{112 B (2\hat{w}_1^{-1} + \hat{w}_0^{-1})(1+\epsilon) }.
\end{align*}
Hence, the right-hand side of (\ref{k}) can be bounded by 
\begin{align}\label{u}
    4 \mathcal{N}_1 (\frac{\epsilon \beta}{5},\mathcal{R}_n,Z_1^n) \exp \left(- \frac{n \epsilon^2 (1-\epsilon) \alpha}{112 B (2\hat{w}_1^{-1} + \hat{w}_0^{-1})(1+\epsilon) }\right).
\end{align}
Combining (\ref{c}), (\ref{d}), (\ref{e}), and (\ref{u}), we have that, for $n > \frac{16B(2 \hat{w}_1^{-1} + \hat{w}_0^{-1})}{\epsilon^2 (\alpha + \beta)} $,
\begin{align*}
    &\mathsf{P} \left(  \exists g \in \mathcal{G}_n : \mathsf{E}[r(Z;g)|Z_1^n] - \frac{1}{n} \sum_{i=1}^{n} r(Z_i;g) \geq \epsilon(\alpha + \beta ) + \epsilon \mathsf{E}[r(Z;g)|Z_1^n] \right) \notag \\
    &\leq \frac{32}{7} \mathcal{N}_1 (\frac{\epsilon \beta}{5},\mathcal{R}_n,Z_1^n) \exp \left(- \frac{n \epsilon^2 (1-\epsilon) \alpha}{112 B (2\hat{w}_1^{-1} + \hat{w}_0^{-1})(1+\epsilon) }\right) \notag\\
    &\hspace{3cm} + \frac{64}{7} \mathcal{N}_1 (\frac{\epsilon (\alpha + \beta)}{5}, \mathcal{R}_n, Z_1^n) \exp \left( -\frac{3 \epsilon^2 (\alpha + \beta )n}{160B^2(2\hat{w}_1^{-1} + \hat{w}_0^{-1})^2}\right) \notag \\
    &\leq 14 \mathcal{N}_1 (\frac{\epsilon \beta}{5},\mathcal{R}_n,Z_1^n) \exp \left( -\frac{\epsilon^2 (1-\epsilon) \alpha n}{54 B^2(2\hat{w}_1^{-1} +\hat{w}_0^{-1})^2 (1+\epsilon)}\right).
\end{align*}
For $n < {16B(2 \hat{w}_1^{-1} + \hat{w}_0^{-1})}/{\epsilon^2 (\alpha + \beta)} $, we have that
\begin{align*}
    \exp \left( -\frac{\epsilon^2 (1-\epsilon) \alpha n}{54 B^2(2\hat{w}_1^{-1} +\hat{w}_0^{-1})^2 (1+\epsilon)}\right) \geq \exp \left( -\frac{2}{135} \right) \geq \frac{1}{14}.
\end{align*}
Hence the assertion follows trivially. \qed

%thm2=========================================================
Now we are ready to prove Theorem \ref{thm2}.

\noindent{\it Proof of Theorem \ref{thm2}:} Let $\alpha = \beta = t/2, \epsilon = 1/2$, then we have 
\begin{align*}%\label{v}
    \mathsf{P} \left( \mathsf{E}r(Z;\hat{g}_n) - \frac{2}{n} \sum_{i=1}^{n} r(Z_i;\hat{g}_n )\geq t \right) \leq \mathsf{P} \left( \exists g \in \mathcal{G}_n : \mathsf{E}r(Z;g) - \frac{2}{n} \sum_{i=1}^{n} r(Z_i;g )\geq t \right).
\end{align*}
By Lemma \ref{lm6}, for $\forall a_n > 0$,
\begin{align*}%\label{w}
    &\mathsf{E}_\mathcal{S} \left[ \mathsf{E}r(Z;\hat{g}_n) - \frac{2}{n} \sum_{i=1}^{n} r(Z_i;\hat{g}_n ) \right] \\ 
    &\leq a_n + \int_{a_n}^\infty \mathsf{P} \left( \mathsf{E}r(Z;\hat{g}_n) - \frac{2}{n} \sum_{i=1}^{n} r(Z_i;\hat{g}_n )\geq t \right) dt \\
                                    &\leq a_n + \int_{a_n}^\infty \mathsf{P} \left( \exists g \in \mathcal{G}_n : \mathsf{E}r(Z;g) - \frac{2}{n} \sum_{i=1}^{n} r(Z_i;g )\geq t \right) dt \\
                                    &\leq a_n + \int_{a_n}^\infty 14 \mathcal{N}_1 (\frac{t}{20},\mathcal{R}_n,Z_1^n) \exp \left( -\frac{nt}{1296B^2(2\hat{w}_1^{-1} + \hat{w}_0^{-1})^2}\right) dt \\
                                    &\leq a_n + 14 \mathcal{N}_1 (\frac{a_n}{20},\mathcal{R}_n,Z_1^n) \exp \left( -\frac{n a_n }{1296B^2(2\hat{w}_1^{-1} + \hat{w}_0^{-1})^2}\right) \cdot \frac{1296B^2(2\hat{w}_1^{-1} + \hat{w}_0^{-1})^2}{n}.
\end{align*}
Choosing $a_n = \log ( 14 \mathcal{N}_1 (a_n/20,\mathcal{R}_n,Z_1^n))\cdot 1296B^2(\hat{w}_1^{-1} + \hat{w}_0^{-1})^2 /n$, note that $a_n / 20 \geq 1/n$, we have 
\begin{align}\label{x}
    \mathsf{E}_\mathcal{S} \left[ \mathsf{E}r(Z;\hat{g}_n) - \frac{2}{n} \sum_{i=1}^{n} r(Z_i;\hat{g}_n ) \right] &\leq \frac{1296B^2(2\hat{w}_1^{-1} + \hat{w}_0^{-1})^2 (\log 14 \mathcal{N}_1 (\frac{1}{n},\mathcal{R}_n,Z_1^n) + 1 )}{n}. 
        % &\leq \frac{1296B^2(2\hat{w}_1^{-1} + \hat{w}_0^{-1})^2 (\log 14 \mathcal{N}_1 (1/(2\hat{w}_1^{-1} + \hat{w}_0^{-1})n , \mathcal{G}_n,Z_1^n) + 1 )}{n}
\end{align}
By Theorem 12.2 of \cite{anthony1999neural}, for $n \geq Pdim(\mathcal{G}_n)/2$, we have
\begin{align}\label{f18}
    \mathcal{N}_1 (1/(2\hat{w}_1^{-1} + \hat{w}_0^{-1})n , \mathcal{G}_n,Z_1^n) \leq \left( \frac{(2\hat{w}_1^{-1} + \hat{w}_0^{-1})eBn^2}{Pdim(\mathcal{G}_n)}\right)^{Pdim(\mathcal{G}_n)}.
\end{align}
By Theorem 10 of \citep{bartlett2019nearly}, there exist some constant $c$ and $C$ such that
\begin{align}\label{f17}
    c\cdot S_n D_n\log S_n/D_n \leq Pdim(\mathcal{G}_n) \leq C\cdot S_n D_n\log S_n.
\end{align}
Combining (\ref{x}), (\ref{f18}), (\ref{f17}), for some constant $c_2 > 0$ we have
\begin{align}\label{f19}
    \mathsf{E}_\mathcal{S} \left[ \mathsf{E}r(Z;\hat{g}_n) - \frac{2}{n} \sum_{i=1}^{n} r(Z_i;\hat{g}_n ) \right] &\leq c_2(2\hat{w}_1^{-1} + \hat{w}_0^{-1})^2 B^2\frac{S_n D_n \log S_n \log n}{n}.
\end{align}
Finally, combine Lemma \ref{lm1}, Lemma \ref{lm2}  and (\ref{f19}), we have 
\begin{align*}
    \mathsf{E}_\mathcal{S} \mathsf{E} r(Z;\hat{g}_n) \leq c_2(2\hat{w}_1^{-1} + \hat{w}_0^{-1})^2 B^2\frac{S_n D_n\log S_n \log n}{n} + 18(2\hat{w}_1^{-1} + \hat{w}_0^{-1})B\beta^2 p^{3\beta/2}(N_n M_n)^{-2\beta/p}.
\end{align*}\qed

%thm3=========================================================
Finally, we utilize Theorem \ref{thm2} to prove Corollary \ref{cor1}.

\noindent{\it Proof of Corollary \ref{cor1}:} 
According to (\ref{f3}), for $g \in \mathcal{G}_n$, we have
\begin{align*}
    \max\{W,D\} \leq S \leq 2W^2D.
\end{align*}
By Theorem \ref{thm2} we have 
\begin{align}\label{f20}
        \mathsf{E}_\mathcal{S} \mathsf{E} r(Z;\hat{g}_n) \leq c_3(2\hat{w}_1^{-1} + \hat{w}_0^{-1})^2 B^2\frac{W_n^2 D_n^2\log W_n^2 D_n \log n}{n} + 18(2\hat{w}_1^{-1} + \hat{w}_0^{-1})B\beta^2 p^{3/2}(N_n M_n)^{-2\beta/p}.
\end{align}
Let $N_n$ be fixed and $W_n = 38\beta^2N_n\lceil\log_2(8N_n)\rceil$. In order to achieve the optimal rate with respect to $n$, we need to balance stochastic error and approximation error. It means we have to find that
\begin{align*}
    \frac{D_n^2\log D_n}{n}\log n \approx M_n^{-2\beta /p}.
\end{align*}
This implies that $M_n = O(n^{d/2(2\beta + p)})$. By plugging in the choice of $M_n$, we have that 
\begin{align*}
    \mathsf{E} r(Z;\hat{g}_n) = O(n^{-\beta / (2\beta + p)}).
\end{align*}
Moreover, using the fact that $\log(1+e^g) - \log(1+e^{g_0}) \leq O(g-g_0) $, we have that $r(Z;\hat{g}_n)$ is of the same order as $g-g_0$. Hence
\begin{align*}
    \mathsf{E}_X |\hat{g}_n (X) - g_0 (X)| = O(n^{-\beta / (2\beta + p)}). 
\end{align*}\qed

\bigskip
	
\noindent {\bf Acknowledgment}

\bibliographystyle{apalike}

\bibliography{bibliography}

\end{document}